%% file: egbib.tex
\documentclass[10pt,twocolumn,letterpaper]{article}

\usepackage[pagenumbers]{cvpr} 

\usepackage{graphicx}
\usepackage{amsmath}
\usepackage{amssymb}
\usepackage{booktabs}
\usepackage{comment}
\usepackage{xcolor}
\usepackage{multirow}

\usepackage{algorithm}
\usepackage{algpseudocode}

\usepackage[pagebackref,breaklinks,colorlinks]{hyperref}

\usepackage[capitalize]{cleveref}
\crefname{section}{Sec.}{Secs.}
\Crefname{section}{Section}{Sections}
\Crefname{table}{Table}{Tables}
\crefname{table}{Tab.}{Tabs.}

\def\cvprPaperID{11400} 
\def\confName{CVPR}
\def\confYear{2022}

\newcommand{\eq}{Eqn.~}%
\newcommand{\textcite}[1]{\cite{#1}}%

\newcommand{\jf}[1]{\textcolor{red}{[\textbf{JF:} #1]}}

\newcommand{\raj}[1]{\textcolor{purple}{[\textbf{RD:} #1]}}

\newcommand{\para}[1]{{\noindent\textbf{#1}}}

\newcommand{\cut}[1]{}

\begin{document}

\title{Regularizing Self-training for Unsupervised Domain Adaptation\\ via Structural Constraints}

\author{Rajshekhar Das$^{1}$\thanks{Equal contribution.}~~\thanks{Correspondence.} ~ Jonathan Francis$^{1,2*}$ ~ Sanket Vaibhav Mehta$^{1*}$ ~ Jean Oh$^{1}$ ~ Emma Strubell$^{1}$ ~ Jos\'{e} Moura$^{1}$\\
$^{1}$Carnegie Mellon University ~ 
$^{2}$Bosch Center for Artificial Intelligence\\
{\tt\small \{rajshekd, jmf1, svmehta, hyaejino,  strubell, moura\}@andrew.cmu.edu}
}
\maketitle

\begin{abstract}
Self-training based on pseudo-labels has emerged as a dominant approach for addressing conditional distribution shifts in unsupervised domain adaptation (UDA) for semantic segmentation problems. A notable drawback, however, is that this family of approaches is susceptible to erroneous pseudo labels that arise from confirmation biases in the source domain and that manifest as nuisance factors in the target domain. A possible source for this mismatch is the reliance on only photometric cues provided by RGB image inputs, which may ultimately lead to sub-optimal adaptation. To mitigate the effect of mismatched pseudo-labels, we propose to incorporate structural cues from auxiliary modalities, such as depth, to regularise conventional self-training objectives. Specifically, we introduce a \textit{contrastive pixel-level objectness constraint} that pulls the pixel representations within a region of an object \textit{instance} closer, while pushing those from different object \textit{categories} apart. To obtain object regions consistent with the true underlying object, we extract information from both depth maps and RGB-images in the form of multimodal clustering. Crucially, the objectness constraint is agnostic to the ground-truth semantic labels and, hence, appropriate for unsupervised domain adaptation. In this work, we show that our regularizer significantly
improves top performing self-training methods (by up to $2$ points) in various UDA benchmarks for semantic segmentation. We include all code in the supplementary.
\end{abstract}

\input{sections/introduction}

\input{sections/relatedwork}

\input{sections/approach}


\input{sections/experiments}


\input{sections/results}


\input{sections/conclusion}

{\small
\bibliographystyle{ieee_fullname}
\bibliography{egbib}
}
\clearpage
\input{sections/supplementary}

\end{document}

%% file: sections/introduction.tex
\section{Introduction}
\label{sec:introduction} 

\begin{figure}[h]
\centering
 \includegraphics[width=\linewidth]{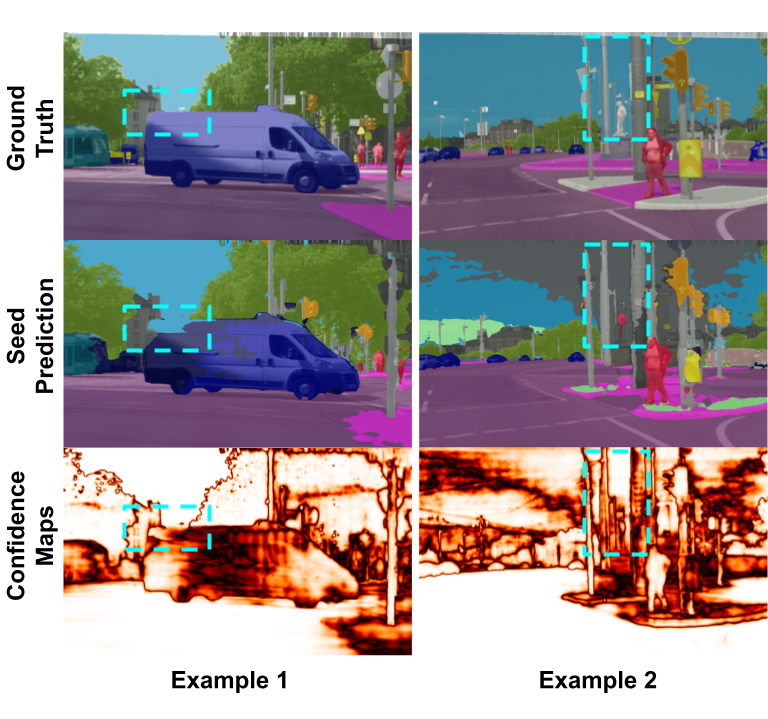}
\caption{\textbf{Motivation for Objectness Constraints:} The above examples compare target-domain ground-truth segmentation, predicted segmentation and prediction confidence (brighter regions are more confident) of a seed model that was adapted from source to target domain via adversarial adaptation \cite{adaptseg2018}. Most self-training approaches use such a seed model to predict pixelwise pseudo-labels. The blue-dashed-boxes highlighte the high-confidence regions that are likely to be included in the set of a pseudo-labels despite being mis-classified. We propose to mitigate the adverse effect of such noisy pseudo-labels on self-training based adaptation via objectness constraints.}
\label{fig:teaser}
\end{figure}

Semantic segmentation is a crucial and challenging task for applications such as autonomous driving \cite{zhang2020transferring,DASAC,vu2019dada,Zhang2019CategoryAU,hoffman2018cycada} that rely on pixel-level semantics of the scene. Performance on this task has significantly improved over the past few years following the advances in deep supervised learning \cite{Chen2018DeepLabSI}. However, an important limitation arises from the excessive cost and time taken to annotate images at a pixel-level (reported to be 1.5 hours per image in a popular dataset \cite{cordts2016cityscapes}). Further, most real-world datasets do not have sufficient coverage over all variations in outdoor scenes such as weather conditions and geography-specific layouts that can be crucial for large-scale deployment of learning-based models in autonomous vehicles. Acquiring training data to cater to such scene variations would significantly add to the cost of annotation.

To address the annotation problem, synthetic datasets curated from 3D simulation environments like GTA \cite{richter2016playing} and SYNTHIA \cite{ros2016synthia} have been proposed where large amounts of annotated data can be easily generated. However, generated data introduces domain shift due to differences in visual characteristics of simulated images (source domain) and real images (target domain). To mitigate such shifts, unsupervised domain adaptation strategies \cite{adaptseg2018,DBLP:journals/corr/BousmalisSDEK16,zou2018unsupervised,Zhang2019CategoryAU,zhang2020transferring,hoffman2018cycada,DASAC} for semantic segmentation have been extensively studied in the recent years. Among these approaches, self-training \cite{NIPS2004_96f2b50b} has emerged as a particularly promising approach that involves pseudo labelling the (unlabelled) target-domain data using a \textit{seed} model trained solely on the source domain. Pseudo-label predictions for which the confidence exceeds a predefined threshold are then used to further train the model and ultimately improve the target-domain performance. 


While self-training based adaptation is quite effective, it is susceptible to erroneous pseudo labels arising from confirmation bias \cite{Arazo2020PseudoLabelingAC} in the seed model. Confirmation bias results from training on source domain semantics that might introduce factors of representation that serve as nuisance factors for the target domain. In the context of semantic segmentation, such a bias manifests as pixel-wise seed predictions that are highly confident but incorrect (see Figure \ref{fig:teaser}).
For instance, if the source domain images usually have bright regions (high intensity of the RGB channels) for the \textit{sky} class, then bright regions in target domain images might be predicted as the sky with high confidence, irrespective of the actual semantic label. Since highly confident predictions qualify as pseudo-labels, training the model on potentially noisy predictions can ultimately lead to sub-optimal performance in the target domain. Thus, in this work, we seek to reduce the heavy reliance of self-training methods on photometric cues for predicting pixel-wise semantic labels. 

To that end, we propose to incorporate auxiliary modality information such as depth maps that can provide structural cues \cite{vu2019dada,Lee2019SPIGANPA, wang2021domain, Chen2019LearningSS}, complementary to the photometric cues. Semantic segmentation datasets are usually accompanied by depth maps that can be easily acquired in practice \cite{cordts2016cityscapes, DEUM}. Since na{\"i}ve fusion of features that are extracted from depth information can also introduce nuisance \cite{Lee2019SPIGANPA, vu2019dada}, an important question is raised --- \textit{How can we leverage the depth modality to counter the effect of noisy pseudo-labels during self-training?} 
In this work, we propose a contrastive \textit{objectness} constraint derived from depth maps and RGB-images in the target domain that is used to regularise conventional self-training methods. The constraint is computed in two steps: an \textit{object-region} estimation step, followed by \textit{pixel-wise contrastive loss} computation. In the first step, we perform unsupervised image segmentation using both depth-based histograms and RGB-images that are fused together to yield multiple object-regions per image. These regions respect actual object boundaries, based on the structural information depth provides, as well as visual similarity. In the second step, the object-regions are leveraged to formulate  a contrastive objective \cite{NIPS2016_6b180037, chen2020simple, NEURIPS2020_d89a66c7} that pulls together  pixel  representations within an object region and pushes apart those from different semantic categories. Such an objective can improve semantic segmentation by causing the pixel  representations of a semantic category to form a compact cluster that is well separated from other categories. We empirically demonstrate the effectiveness of
our constraint on popular benchmark tasks, \texttt{GTA$\rightarrow$Cityscapes} and \texttt{SYNTHIA$\rightarrow$Cityscapes}, on which we achieve competitive segmentation performance. To summarise our contributions:
 
\begin{itemize}
\item We propose a novel objectness constraint derived from depth and RGB information to regularise self-training approaches in unsupervised domain adaptation for semantic segmentation. The use of multiple modalities introduces \textit{implicit} model supervision that is complementary to the pseudo-labels and hence, lead to a more robust self-training.
\item We empirically validate the most important aspect of our regulariser, \ie, its ability to improve a variety of self-training methods. Specifically, our approach achieves $1.2\%$-$2.9\%$ (GTA) and $2.2\%$-$4.4\%$ (SYNTHIA) relative improvements over three different self-training baselines. Interestingly, we observe that regularisation improves performance on both ``stuff'' and ``things'' classes, somewhat normalising  the effects of classwise statistics.
\item Further, our regularised self-training method achieves state-of-the-art mIoU of $54.2\%$ in \texttt{GTA $\rightarrow$ Cityscapes} settings and improves classwise IoUs by up to $4.8\%$ over best prior results. 
\end{itemize}

%% file: sections/relatedwork.tex
\section{Related Work}
\label{sec:related_work}




\para{Unsupervised domain adaptation.} Unsupervised domain adaptation (UDA) is of particular importance in complex structured-prediction problems, such as semantic segmentation in autonomous driving, where the domain gap between a source domain (e.g., an urban driving dataset) and target domain (real-world driving scenarios) can have devastating consequences on the efficacy of deployed models. Several approaches \cite{Ganin_2017,DBLP:journals/corr/BousmalisSDEK16,Pan2020UnsupervisedIA,hoffman2018cycada, 9156459} have been proposed for learning domain invariant representations, e.g., through adversarial feature alignment \cite{Ganin_2017, DBLP:journals/corr/BousmalisTSKE16, 8099799, Wang2020DifferentialTF}, which addresses the domain gap by minimising a distance metric that characterises the divergence between the two domains \cite{5640675, DBLP:conf/icml/LongC0J15, Long2017DeepTL, 6751205, 10.5555/3305890.3305999, 10.5555/3294996.3295130, 4967588, Muandet13DG}. Problematically, such approaches address only shifts in the marginal distribution of the covariates or the labels and, therefore, prove insufficient for handling the more complex shifts in the conditionals \cite{Johansson2019SupportAI, DBLP:journals/corr/abs-1901-09453, Wu2019DomainAW}. Self-training approaches have been proposed to induce category-awareness \cite{Zhang2019CategoryAU} or cluster density-based assumptions \cite{shu2018dirt}, in order to anchor or regularise conditional shift adaptation, respectively. In this paper, we build upon these works by jointly introducing category-awareness through the use of pseudo-labeling strategies and regularisation through the definition of contrastive depth-based objectness constraints.

\para{Self-training with pseudo-labels.} Application of self-training has become popular in the sphere of domain adaptation for semantic segmentation \cite{zou2018unsupervised, li2019bidirectional, Zhang2019CategoryAU, TextIR}. Here, pseudo-labels are assigned to observations from the target domain, based on the semantic classes of high-confidence (e.g., the closest or least-contrastive) category centroids \cite{Zhang2019CategoryAU, xie2018learning}, prototypes \cite{chen2019progressive}, cluster centers \cite{kang2019contrastive}, or superpixel representations \cite{zhang2020transferring} that are learned by a model trained on the source domain. Often, to ensure the reliability of initial pseudo labels for target domain, the model is first warmed up via adversarial adaptation \cite{Zhang2019CategoryAU,zhang2020transferring}. Moreover, for stability purposes, pseudo labels are updated in a stagewise fashion, thus resulting in an overall complex adaptation scheme. Towards streamlining this complex adaptation process, recent approaches like \cite{DASAC, Tranheden2021DACSDA} propose to train without adversarial warmup and with a momentum network to circumvent stagewise training issue. A common factor underlying most self-training methods is their reliance on just RGB inputs that may not provide sufficient signal for predicting robust target-domain pseudo labels. This motivates us to look for alternate forms of input like depth that is easily accessible and provide a more robust signal.

\para{Adaptation with multiple modalities.} Learning and adaptation using multimodal contexts presents an opportunity for leveraging complementarity between different views of the input space, to improve model robustness and generalisability. In the context of unsupervised domain adaptation, use of mutimodal information has recently become more popular with pioneering works like \cite{Lee2019SPIGANPA}. Specifically, \cite{Lee2019SPIGANPA} uses depth regression as a way to regularise the GAN based domain translation resulting in better capture of source semantics in the generated target images. Another related approach
\cite{vu2019dada} proposes the use of depth via an auxiliary objective to learn features that when fused with  primary semantic segmentation prediction branch provides a more robust representation for adaptation. While sharing our motivation for use of auxiliary information, their use of fused features for adaptation does not address the susceptibility of adversarial adaptation to conditional distribution shifts. In contrast to this method, we propose a depth based objectness constraint for adaptation via self-training that not only leverages multimodal context but also handles conditional shifts more effectively. Moreover, unlike the previous works that use depth only for the source domain, we explore its application exclusively to the target domain. Contemporary to our setting, \cite{wang2021domain} improves adaptation by extracting the correlation between depth and RGB in both domains. An important distinction of our approach with regards to above works is that we exploit  the complementarity of RGB and depth instead of the correlation to formulate a contrastive regularizer.  The importance of multimodal information has also been considered in other contexts such as indoor semantic segmentation \cite{stekovic2020casting} and adaptation for 3D segmentation using 2D images and 3D points clouds \cite{xmuda}. While not directly related to our experimental settings, they provide insight and inspiration for our approach.

%% file: sections/approach.tex
\begin{figure*}[h]
\centering
\includegraphics[width=\linewidth]{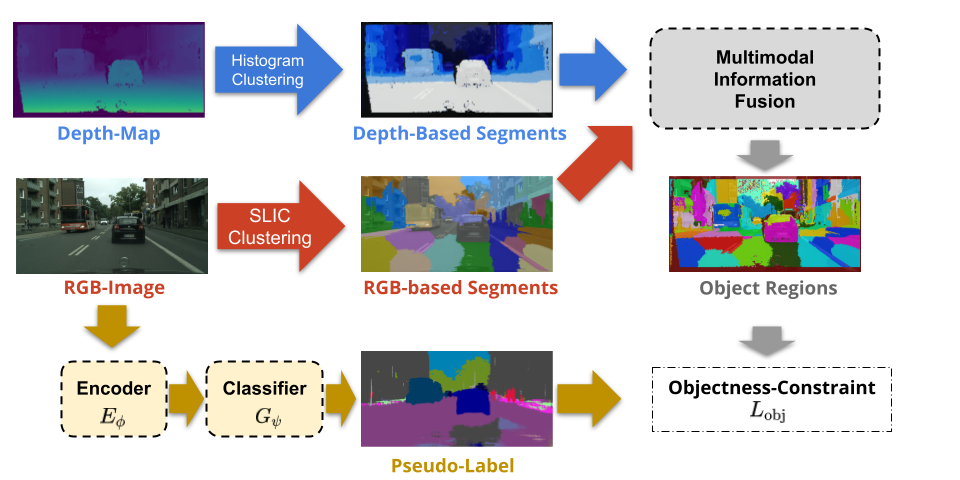}
\caption{\textbf{Objectness Constraint Formulation:} Overall pipeline for computing the objectness constraint using multi-modal object-region estimates derived from RGB-Image and Depth-Map. Depth segmentation is obtained by clustering the histogram of depth values and  RGB segmentation is obtained via k-means clustering (SLIC) of raw-pixel intensities.  
Fusing these two types of segmentation yields object regions that are more consistent with the actual object. For example, a portion of the car in the middle is wrongly clustered with the road in depth segmentation and with the left-wall under RGB segmentation. However, the fused segmentation yields car-regions that completely respect the boundary of the car.} 
\label{fig:pipeline}
\end{figure*}

\section{Self-Training with Objectness Constraints}
\label{sec:approach}

We begin by introducing preliminary concepts on self-training based adaptation. These concepts serve as bases for introducing our objectness constraint in Section \ref{sec:approach} that is used to regularise the self-training methods. We refer to our framework as \texttt{PAC-UDA} which uses \textbf{P}seudo-labels \textbf{A}nd objectness \textbf{C}onstraints for self-training in \textbf{U}nsupervised \textbf{D}omain \textbf{A}daptation for semantic segmentation. Although, we describe a canonical form of self-training for formalising our regularisation constraint, \texttt{PAC-UDA} should be seen as a general approach that can encompass various forms of self-training (as shown in experiments).

\para{Unsupervised Domain Adaptation (UDA) for Semantic Segmentation:} Consider a dataset $D^s = \{(x_i^s, y_i^s)\}_{i=1}^{N_s}$ of input-label pairs sampled from a source domain distribution, $P_{X\times Y}^s$. The input and labels share the same spatial dimensions, $H \times W$,  where each pixel of the label is assigned a class $c\in \{1, \ldots, C\}$ and is represented via a $C$ dimensional one-hot encoding. We also have a dataset $D^t = \{(x_i^t, y_i^t)\}_{i=1}^{N_t}$ sampled from a target distribution, $P_{X\times Y}^t$ where the corresponding labels, $\{y_i^t\}$ are \textit{unobserved} during training. Here, the target domain is separated from the source domain due to domain shift expressed as  $P_{X\times Y}^s \neq P_{X\times Y}^t$. Under such a shift, the goal of unsupervised domain adaptation is to leverage $D^s$ and $D^t$ to learn a parametric model that performs well in the target domain. The model is defined as a composition of an encoder, $E_\phi:X \to \mathcal{Z}$ and a classifier, $G_\psi: \mathcal{Z} \to \mathcal{Z}_P $ where, $\mathcal{Z}\in \mathbb{R}^{H \times W \times d}$ represents the space of  $d$-dimensional spatial embeddings, $\mathcal{Z}_P\in \mathbb{R}^{H \times W \times C}$ gives the un-normalized distribution over the $C$ classes at each spatial location, and $\{\phi, \psi\}$ are the model parameters. To learn a suitable target model, the parameters are  optimised using a cross-entropy objective on the source domain, 
\begin{align}
L_\text{cls}^s &= -\frac{1}{N_s}\sum\limits_{i=1}^{N_s}\sum\limits_{m=1}^{H\times W}\sum\limits_{c=1}^{C}y_{imc}^s\log p^s_{imc}(\psi, \phi) \\
p^s_{imc}(\psi, \phi)  &= \sigma \left(G_\psi \circ E_\phi(x_i^s)\right)|_{m,c}~,
\label{eq:CEsrc}
\end{align}
\noindent where $\sigma$ denotes softmax operation and an adaptation objective over the target domain as described next.
 
\para{Pseudo-label self-training (PLST):} Following prior works \cite{zou2018unsupervised, Zhang2019CategoryAU}, we describe a simple and effective approach to PLST that leverages a source trained seed model to \textit{pseudo}-label unlabelled target data, via confidence thresholding.
Specifically, the seed model is first trained on $D^s$ using \eq\ref{eq:CEsrc} to obtain a good parameter initialisation, $\{\phi_0, \psi_0\}$. Then, this model is used to compute pixel-wise class probabilities, $p^t_{im}(\psi_0, \phi_0)$  using to \eq\ref{eq:CEsrc} for each target image, $x_i^t \in D^t$. These probabilities are used in conjunction with a predefined threshold $\delta$, to obtain one-hot encoded pseudo-labels
\begin{equation}
\tilde{y}_{imc}^t = 
\begin{cases}
1 ~\text{if}~ c=\arg\max\limits_{c'} p^t_{imc'} ~\text{and}~ p^t_{imc} \geq \delta\\
0 ~\text{otherwise} 
\end{cases} 
\label{eq:pslab}
\end{equation} 
Note that while \eq \ref{eq:pslab} uses a class-agnostic fixed 
threshold in practice, this threshold can be made class-specific and dynamically updated over the course of self-training.
Such a threshold ensures that only the highly-confident predictions contribute to successive training. The final self-training objective can be written in terms of pseudo-labels as  
\begin{equation}
L_\text{st}^t = -\frac{1}{N_t}\sum\limits_{i=1}^{N_t}\sum\limits_{m=1}^{H\times W}\sum\limits_{c'=1}^{C}\tilde{y}_{imc'}^t\log \left(p^t_{imc'}\right)
\label{eq:st}
\end{equation}
The overall UDA objective is simply, $L_\text{uda} = L_\text{cls}^s + \alpha_\text{st}  L_\text{st}^t$, where $\alpha_\text{st}$ is the relative weighting coefficients. 

\subsection{Supervision For Objectness Constraint}
\label{sec:superv}


An important issue with the self-training scheme described above is that it is usually prone to confirmation bias  that can lead to compounding errors in target model predictions when trained on noisy pseudo-labels.
To alleviate target performance, we introduce auxiliary modality information (like, depth) that can provide indirect supervision for semantic labels in the target domain and improve the robustness of self-training. In this section we describe our multimodal objectness constraint that  extracts object-region estimates to formulate a contrastive objective. The overview of our objectness constraint formulation is presented in  Fig. \ref{fig:pipeline}.
 
\para{Supervision via Depth:} Segmentation datasets are often accompanied with depth maps registered with the RGB images. In practice, depth maps can be obtained from stereo pairs \cite{cordts2016cityscapes, DEUM} or sequence of images \cite{Godard2019DiggingIS}. These depth maps can reveal the presence of distinct objects in a scene. We particularly seek to extract object regions from these depth maps by first computing a histogram of depth values with predefined, $b$ number of bins. We then leverage the property of objects under "things" categories \cite{Hariharan2014SimultaneousDA}  whose range of the depth is usually much smaller than the range of entire scene depth. Examples of such categories in outdoor scene segmentation include persons, cars, poles etc. This property translates into high density regions (or peaks) in the histogram corresponding to distinct objects at distinct depths. Among these peaks, we use the ones with prominence \cite{TP} above a threshold, $\delta_\text{peak}$ as centers to cluster the histograms into discrete regions with unique labels. These labels are then assigned to every pixel whose depth values lie in the associated region. An example of the resulting depth-based segmentation for $b=200$ and $\delta_\text{peak}=0.0025$ is visualised in Fig. \ref{fig:pipeline}.

\para{Supervision via RGB:} Another important form of self-supervision for object region estimates is based on RGB-input clustering. We adopt SLIC \cite{6205760} as a fast algorithm for partitioning images into multiple segments that respect object boundaries; the SLIC method applies k-means clustering in pixel space to group together adjacent pixels that are visually similar.
An important design decision is the number of SLIC segments, $k_s$: small $k_s$ leads to large cluster sizes that is agnostic to the variation in object scales, across different object categories and instances of the scene. Consequently, pixels from distinct object instances may be grouped together regardless of the semantic class, thus violating the notion of object region. Conversely, a large $k_s$ will over-segment each object in the scene, resulting in a trivial objectness constraint. Triviality arises from enforcing similarity of pixel-embeddings that share roughly identical pixel neighbourhoods and hence are likely to yield the same class predictions anyway. 

Thus, to formulate a non-trivial constraint with sufficiently small $k_s$ that also respects object boundaries, we propose to fuse region estimates from both depth and RGB modalities.We first obtain $k_s$ segments using SLIC over the RGB image followed by further partitioning of each segment into smaller ones based on the depth segmentation. The process, visualised in Fig. \ref{fig:pipeline} highlights the importance of our multimodal approach. Purely depth based segments are agnostic to pixel intensities and  may cluster together distinct object categories that lie at similar depths, for instance, the car in the front and the sidewalk. On the other hand, purely RGB segments with sufficiently small $k_s$ may assign the same cluster label even to objects at distinct depths, for example, the back of the bus and the small car at the back. In contrast, object regions derived from a fusion of these two modalities can lead to object regions that are more consistent with individual object instances (for example, the small car at the back as well as the car in the front). We empirically demonstrate the effectiveness of objectness constraint derived from such multimodal fusion in Section \ref{sec:ablate}.

\input{sections/tab_generality}

\subsection{Objectness Constraints through Contrast}
\label{subsec:constraints} 
Our objectness constraint is formulated using a contrastive objective that pulls together pixel representations within an object region and pushes apart those that belong to different object categories. Formally, we assign a region \textit{index} and a region \textit{label} to every pixel associated with an object region of the input scene. Each region index is a unique natural number in $\{1, \ldots, K\}$ where $K$ is the number of object regions. A region label is assigned as the most frequent pseudo-label class within the object region. In practice, noisy pseudo-labels can lead to region labelling that is inconsistent with true semantic labels. To minimise such inconsistencies, we introduce a threshold $\tau_p$ that selects \textit{valid} object regions for which the proportion of pixels with pseudo-label class same as the region label is above this threshold. This selection excludes the object regions with no dominant pseudo-label class from contributing to the objectness constraint. Since the cost of computing pairwise constraints is quadratic in the number of pixels, we recast the pairwise constraint into a protoypical loss that reduces the time complexity to linear. Towards the end, we first compute a prototypical representation for each region using the associated pixel embeddings, 
\begin{align}
\nu_{k} = \frac{1}{|U_k|} \sum_{{p\in U_k}}z_{p}
\end{align}
where, $U_k$ is the set of pixel locations with the $k^\text{th}$ object-region. Then a similarity score (based on Cosine metric) is computed between each pixel and prototypical representation that forms the basis for our contrastive objectness constraint as 
\begin{align}
L_\text{obj}^t &=  \frac{1}{S}\sum\limits_{k}\sum\limits_{p\in U_k}L_\text{obj}^t(p) \\
L_\text{obj}^t(p) &=-\log\left({\frac{\exp(\tilde{z}_p\cdot \tilde{\nu}_{k})}{\sum\limits_{k' \in \Omega(k)}\exp(\tilde{z}_p\cdot \tilde{\nu}_{k'})}}\right)
\label{eq:reguda}
\end{align}where, $S$ is the total number of valid pixels, $\Omega(k)$ is the set of valid object regions that have region labels other than $k$, and $\tilde{z}_p$ and $\tilde{\nu}_k$ represent $L_2$ normalised embeddings. Note that the objectness constraints are only computed for the target domain images since we are interested in improving target domain performance using self-training. Additionally, the constraint in \eq~\ref{eq:reguda} is defined for a single image but can be easily extended to multiple images by simply averaging over them; the final regularised self-training objective is then defined as $L_\text{pac} = L_\text{uda}^s  + \alpha_\text{obj} * L_\text{obj}^t $, where $\alpha_\text{obj}$ controls the effect of the constraint on overall training. 

\subsection{Learning and Optimization}
\label{subsec:learning}
To train the our model, PAC-UDA with a base self-training approach, we follow the exact procedure outlined by the corresponding approach. The only difference is that we plug in our constraint as a regularise to the base objective, $L_\text{uda}$. One important consideration is that our regularise depends on reasonable quality of pseudo labels to define region labels that are not random. Thus the regularisation weight, $\alpha_\text{obj}$ is set to zero for a few initial training iterations, post which it switches to the actual value.

%% file: sections/tab_generality.tex
\begin{table*}[h]
 \caption{\footnotesize\textbf{Test of Generality:} We compare the performance of regularised and un-regularised versions of \textit{three} self-training approaches for two domain settings, namely, \texttt{GTA $\rightarrow$ Cityscapes} and \texttt{SYNTHIA $\rightarrow$ Cityscapes}. Both per-class IoU and mean IoUs are presented. The numbers in \textbf{bold} indicate higher accuracies in the pairwise comparisons, between a base-method and the base-method$+$PAC.}
\label{tab:generality}
  \centering
  \scalebox{0.645}{
  \begin{tabular}{ccccccccccccccccccccccc}
    \toprule
  %
  Source Domain & Method &\rotatebox{90}{road}& \rotatebox{90}{sidewalk}& \rotatebox{90}{building} &\rotatebox{90}{wall} &\rotatebox{90}{fence} &\rotatebox{90}{pole} &\rotatebox{90}{light} & \rotatebox{90}{sign} &\rotatebox{90}{vege.} &\rotatebox{90}{terrain} &\rotatebox{90}{sky} &\rotatebox{90}{person} &\rotatebox{90}{rider} &\rotatebox{90}{car} &\rotatebox{90}{truck} &\rotatebox{90}{bus} &\rotatebox{90}{train} &\rotatebox{90}{motor} &\rotatebox{90}{bike} &mIoU \\
\midrule
\multirow{6}{*}{\rotatebox{90}{GTA}}& CAG \cite{Zhang2019CategoryAU}&87.0&44.6&82.9&32.1&35.7&40.6&38.9&45.5&82.6&23.5&78.7&64.0&27.2&84.4&17.5&34.8&35.8&26.7&32.8& 48.2 \\
& CAG $+$ PAC (ours)&86.3&45.7&84.5&30.5&35.5&38.9&40.3&49.9&86.0&33.5&81.1&64.1&25.5&84.5&21.3&32.9&36.3&26.7&40.0& \bfseries49.6\\
\cmidrule(r){2-22}
& SAC\cite{DASAC} & 89.9 &54.0&86.2&37.8&28.9& 45.9&46.9&47.7&88.0&44.8&85.5&66.4&30.3&88.6&50.5&54.5&1.5&17.0&39.3&52.8\\
& SAC $+$ PAC (ours) &93.3&  63.6&  87.2&42.0&25.4&44.9&49.0&50.6&88.1&45.2&87.6&64.0&28.1&83.6&37.5&43.9&13.7&20.1&46.2& \bfseries53.4\\ \cmidrule(r){2-22}
& DACS\cite{Tranheden2021DACSDA}& 93.4&54.3&86.3&28.6&33.7&37.0&41.1&50.6&86.1&42.6&87.6&63.5&28.9&88.1&44.2&52.7&1.7&34.7&48.1&52.8 \\
& DACS $+$ PAC (ours) &93.2&58.8&87.2&33.3&35.1&38.6&41.8&51.4&87.4&45.8&88.3&64.8&31.6&84.3&51.7&53.4&0.6&31.3&50.6&\bfseries54.2\\
\midrule
\midrule
\multirow{6}{*}{\rotatebox{90}{SYNTHIA}}& CAG &87.0&41.0&79.0&9.0&1.0&34.0&15.0&11.0&81.0&-&81.0&55.0&16.0&77.0&-&17.0&-&2.0&47.0&40.8\\
& CAG $+$ PAC (ours)&87.0&42.0&80.0&12.0&3.0&30.0&17.0&17.0&80.0&-&88.0&57.0&5.0&75.0&-&20.0&-&1.0&52.0&\bfseries41.7\\
\cmidrule(r){2-22}
& SAC \cite{DASAC}& 91.7&52.7&85.1&22.6&1.5&42.2&44.1&30.9&82.5&-&73.8&63.0&20.9&84.9&-&29.5&-&26.9&52.2&50.3\\
& SAC $+$ PAC (ours) &83.2&40.5&85.4&30.0&2.0&43.0&42.2&33.8&86.3&-&89.8&65.3&33.5&85.1&-&35.2&-&29.9&55.3&\bfseries52.5\\
\cmidrule(r){2-22}
& DACS \cite{Tranheden2021DACSDA} &84.9&23.0&83.7&16.0&1.0&36.3&35.0&42.8&81.7&-&89.5&63.5&34.5&85.3&-&41.5&-&31.2&50.8&50.0\\
& DACS $+$ PAC (ours) & 90.6&46.7&83.3&18.7&1.3&35.1&34.5&32.0&85.1&-&88.5&66.0&35.0&83.8&-&43.1&-&28.8&46.7&\bfseries51.2\\
\bottomrule
 \end{tabular}}
\end{table*}

%% file: sections/experiments.tex
\section{Experiments}
\label{sec:experiments}

\para{Datasets and Evaluation Metric:} We evaluate the PAC-UDA framework in two common scenarios: the \texttt{GTA\cite{richter2016playing}$\rightarrow$Cityscapes} \cite{cordts2016cityscapes} transfer semantic segmentation task and the \texttt{SYNTHIA\cite{ros2016synthia}$\rightarrow$Cityscapes} \cite{cordts2016cityscapes} task. GTA5 is composed of $24,966$ synthetic images with resolution $1914\times 1052$ and has annotations for $19$ classes that are compatible with the categories in Cityscapes. Similarly, SYNTHIA consists of $9,400$ synthetic images of urban scenes at resolution $1280 \times 760$ with annotations for only $16$ common categories. Cityscapes has of $5,000$ real images and aligned depth maps of urban scenes at resolution 2048 × 1024 and is split into three sets of $2,975$ train, $500$ validation and $1,525$ test images. Of the  $2,975$, we use $2,475$ randomly selected images for self-training and remaining $500$ images for validation. We report the final test performance of our method on the $500$ images of the official validation split. The data-splits are consistent with prior works \cite{DASAC,zhang2020transferring}. The performance metrics used are per class Intersection over Union (IoU) and mean IoU (mIoU) over all the classes. 

\para{Implementation Details:} For object region estimates, we experiment with three different numbers, $k_s \in \{25, 50, 100\}$ of RGB-clusters, two values of prominence thresholds, $\delta_\text{peak} \in \{0.001, 0.0025\}$ and three numbers of histogram bins, $b \in \{100,200,400\}$. Depth maps obtained from stereo pairs can have missing values at pixel-level, as is the case with Cityscapes. These missing values have a value zero and are ignored while generating depth segments using depth-histogram. Finally, due to high computational cost of computing the contrastive objective from pixel-wise embedding, we set the spatial resolution of these embeddings to $256 \times 470$ in CAG and SAC and $300\times 300$ in DACS. We fixed the relative weighting of the regularizer,  $\alpha_\text{obj}$ to $1.0$ as the target performance was found to be insensitive to the exact value. For hyperameter choices regarding architecture and optimizers, we exactly follow the respective self-training base methods \cite{Zhang2019CategoryAU,DASAC,Tranheden2021DACSDA}. Experiments were conducted on $4\times11$GB RTX 2080 Ti GPUs with PyTorch implementation. Further details in the supplementary.

%% file: sections/results.tex
\input{sections/tab_gta2cscape}
\input{sections/tab_syn2cscape}

\subsection{Generality of Objectness Constraint}
In  Table \ref{tab:generality}, we test the generality of our proposed regularizer on three base methods, namely, CAG \cite{Zhang2019CategoryAU}, SAC \cite{DASAC} and DACS \cite{Tranheden2021DACSDA} that generate pseudo labels in different ways.
We use official implementations of each base method with almost same configurations for data preprocessing, model architecture, and optimizer except for a few modifications as follows. In the case of CAG, we replace the Euclidean metric with a Cosine metric as it was found to generate more reliable pseudo-labels. Also, we run it for a single self-training iteration instead of three\cite{Zhang2019CategoryAU}. For the SAC method, we reduce the \texttt{GROUP\_SIZE} from default value of $4$ to $2$ following GPU constraints. Finally, for the DACS approach, we adopt the training and validation splits of Cityscapes used in SAC to maintain benchmark consistency across different base methods. In terms of architecture, DACS and SAC use a standard DeepLabv2 \cite{Chen2018DeepLabSI} backbone whereas CAG augments this backbone with a decoder model (see \cite{Zhang2019CategoryAU} for details). For the sake of fair comparison, we try our best to achieve baseline accuracies that are at least as good as  the published results. While we achieved slightly lower performance on SAC due to resource constraints, we achieve superior accuracies for DACS and CAG baselines. Thus, these methods serve as strong baselines for evaluating our approach.

\begin{figure*}[h]
\centering
 \includegraphics[width=\linewidth]{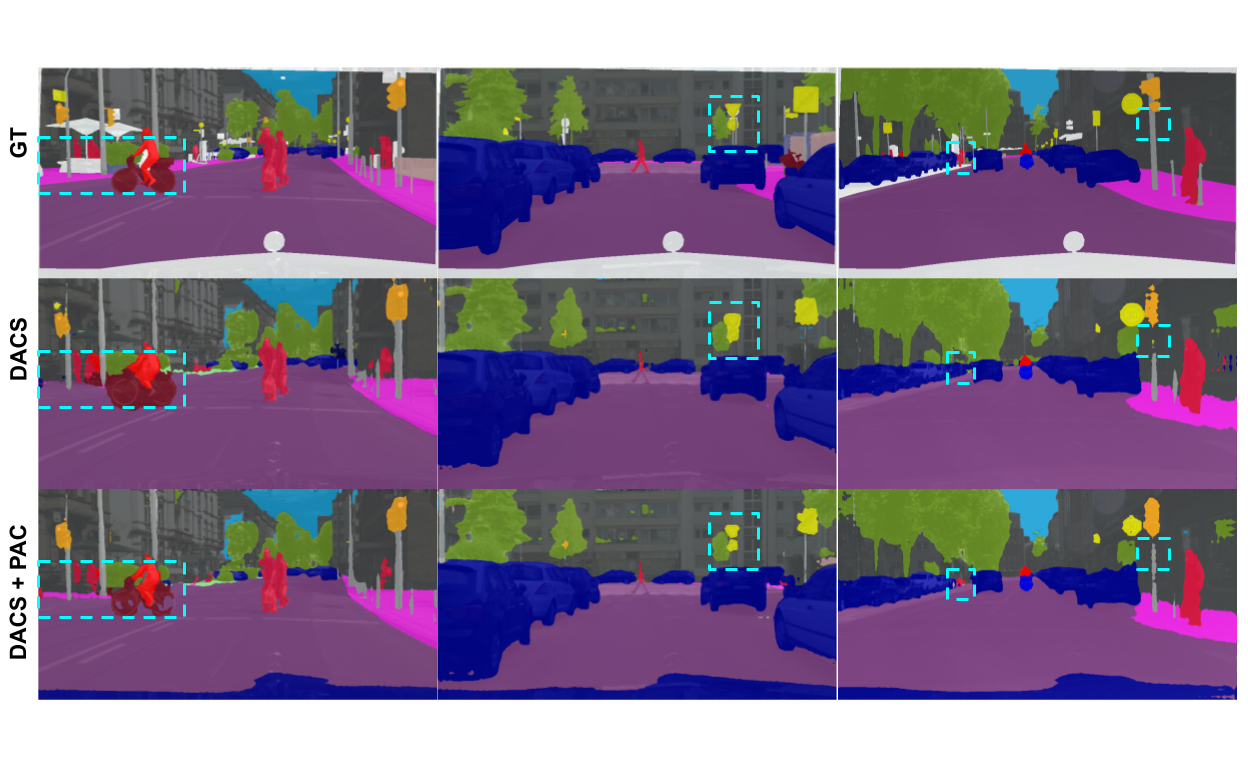}
\caption{\textbf{Qualitative results on Cityscapes\cite{cordts2016cityscapes} post adaptation from GTA\cite{richter2016playing}:} Blue dashed boxes highlight the semantic classes that our regularized version (DACS+PAC) is able to predict more accurately than the base method (DACS). Further visualisations are provided in the supplementary.}
\label{fig:qual}
\end{figure*}

From Table \ref{tab:generality}, we observe that base methods regularised with our constraint always, and sometimes significantly, outperforms the unregularised version in terms of mIoU (by up to $2.2\%$). Secondly, the improvement is across various categories of both \textit{stuffs}  and \textit{things} type. Some of these include sidewalk (up to $9.6\%$), sky (up to $2.4\%$), traffic light (up to $2.1\%$), traffic sign  (up to $4.4\%$) and bike (up to $7.2\%$) classes under  \texttt{GTA$\rightarrow$Cityscapes} while wall (up to $7.4\%$), fence (up to $2.0\%$), person (up to $2.5\%$) and bus (up to $5.7\%$) classes under \texttt{SYNTHIA$\rightarrow$Cityscapes}. While different adaptation settings favour different classes, a particularly striking observation is that large gains are obtained in both frequent (sidewalk, wall) and less-frequent (bus, bike) classes. We suspect that such uniformity arises from our object-region aware constraint that is agnostic to the statistical dominance of specific classes. Finally, Fig. \ref{fig:qual} visualises these observations by comparing the predictions of DACS and DACS+PAC models (trained on GTA) on randomly selected examples from Cityscapes validation split.

\input{sections/tab_ablations}

\subsection{Prior Works Comparison}
\label{sec:results}
In this section, we compare our best performing method with prior works under each domain settings

\para{\texttt{GTA$\rightarrow$Cityscapes} (Table \ref{tab:g2c}):}  In terms of mIoU, our DACS+PAC outperforms the state-of-the-art (SAC) by $0.4\%$ despite having a simpler training objective (no focal loss regularizer or importance sampling) and no adaptive batch normalisation. In particular, our approach outperforms SAC significantly in road, sidewalk, fence,terrain, sky, rider, motorcycle and bike classes by $ 1.9\% - 11.3\%$.
More interestingly, this observation holds when compared to other prior works as well,  wherein our  model improves IoUs for both dominant categories like road and sidewalk as well as less frequent categories like traffic-sign and terrain. For classes like sidewalk, we suspect that structural constraints based on our regularizer reduces contextual bias \cite{Shetty_2019_CVPR}, responsible for coarse boundaries.

\para{\texttt{SYNTHIA$\rightarrow$Cityscapes} (Table \ref{tab:s2c}):} In this setting, our best performing method outperforms all but one prior methods, often by significant margins. While our SAC+PAC under resource constraints compares favourably to the official implementation of SAC (with larger \texttt{GROUP\_SIZE}), it significantly outperforms our implementation of SAC which is a more fair comparison due to same resource constraints. Nevertheless, our approach improves  the best previous results on \textit{wall} class  by $3.5\%$ and achieves state-of-the-art on \textit{pole} and \textit{sign} classes.  

\subsection{Ablations}
\label{sec:ablate}
In this section,  we deconstruct our multi-modal regularizer (PAC) to quantify the effect of individual components on final performance. In Table \ref{tab:ablate}, the `ALL' configuration corresponds to our original formulation. `Only PL' configuration estimates the object-regions using just the pseudo-labels and hence ignores complementary information from depth. `Only Depth+RGB segments' do not use pseudo-labels to define region labels and instead treats each Depth+RGB segment as a unique object category. The configurations in next two rows use only one of the two modalities for estimating object regions while still using pseudo-labels to define region labels.  
We observe that contrastive regulariser based on only 
 pseudo-labels performs the worst and significantly below the one based on just multimodal segments. This is intuitive because reusing pseudo-labels as a regularisation without auxiliary information reinforces the confirmation bias. While, purely RGB based segments lead to better objectness constraint than purely depth-based ones (as can be seen in Fig. \ref{fig:pipeline}), combining the two (ALL config.) yields the best results.  

%% file: sections/tab_gta2cscape.tex
\begin{table*}[h]
 \caption{\footnotesize\textbf{\texttt{GTA $\rightarrow$ Cityscapes} results:} Classwise and mean (over 16 classes) IoU comparison of our DACS$+$PAC with prior works. $^\dagger$ denotes the use of PSPNet \cite{ZhaoSQWJ17}, * denotes  our implementation of SAC with a restricted configuration (\texttt{GROUP\_SIZE}=2) compared to original SAC method (\texttt{GROUP\_SIZE}=4). All other methods use DeepLabV2\cite{Chen2018DeepLabSI} architecture.}
\label{tab:g2c}
  \centering
  \scalebox{0.71}{
  \begin{tabular}{cccccccccccccccccccccc}
    \toprule
      & \rotatebox{90}{road}& \rotatebox{90}{sidewalk}& \rotatebox{90}{building} &\rotatebox{90}{wall} &\rotatebox{90}{fence} &\rotatebox{90}{pole} &\rotatebox{90}{light} & \rotatebox{90}{sign} &\rotatebox{90}{vege.} &\rotatebox{90}{terrain} &\rotatebox{90}{sky} &\rotatebox{90}{person} &\rotatebox{90}{rider} &\rotatebox{90}{car} &\rotatebox{90}{truck} &\rotatebox{90}{bus} &\rotatebox{90}{train} &\rotatebox{90}{motor} &\rotatebox{90}{bike} &mIoU \\
    \midrule
      AdvEnt \cite{vu2018advent}& 89.4& 33.1& 81.0& 26.6& 26.8& 27.2& 33.5& 24.7& 83.9& 36.7& 78.8& 58.7& 30.5& 84.8& 38.5& 44.5& 1.7& 31.6& 32.4& 45.5\\    
    DISE \cite{chang2019adapting}& 91.5& 47.5& 82.5& 31.3& 25.6& 33.0& 33.7& 25.8& 82.7& 28.8& 82.7& 62.4& 30.8& 85.2& 27.7& 34.5& 6.4& 25.2& 24.4& 45.4\\
    Cycada \cite{hoffman2018cycada} & 86.7& 35.6& 80.1& 19.8& 17.5& 38.0& 39.9& 41.5& 82.7& 27.9& 73.6& 64.9& 19.0& 65.0& 12.0& 28.6& 4.5& 31.1& 42.0& 42.7\\
      BLF \cite{li2019bidirectional}& 91.0& 44.7& 84.2& 34.6& 27.6& 30.2& 36.0& 36.0& 85.0& 43.6& 83.0& 58.6& 31.6& 83.3& 35.3& 49.7& 3.3& 28.8& 35.6& 48.5 \\
    CAG-UDA \cite{Zhang2019CategoryAU}& 90.4& 51.6& 83.8& 34.2& 27.8& 38.4& 25.3& 48.4& 85.4& 38.2& 78.1& 58.6& 34.6& 84.7& 21.9& 42.7& \bfseries41.1& 29.3& 37.2& 50.2 \\ 
    PyCDA$^\dagger$ \cite{LianDLG19} & 90.5 & 36.3 & 84.4 & 32.4& 28.7 & 34.6 & 36.4 & 31.5 & 86.8 & 37.9 & 78.5 & 62.3 & 21.5 & 85.6 & 27.9 & 34.8 & 18.0 & 22.9 & 49.3 & 47.4 \\
CD-AM \cite{Yang_2021_WACV} & 91.3 & 46.0 & 84.5 & 34.4 & 29.7 & 32.6 & 35.8 & 36.4 & 84.5 & 43.2 & 83.0 & 60.0 & 32.2 & 83.2 & 35.0 & 46.7 & 0.0 & 33.7 & 42.2 & 49.2 \\
FADA \cite{Wang_2020_ECCV} & 92.5 & 47.5 & 85.1 & 37.6 &  32.8 & 33.4 & 33.8 & 18.4 & 85.3 & 37.7 & 83.5 & 63.2 & \bfseries 39.7 & 87.5 & 32.9 & 47.8 & 1.6 & \bfseries34.9 & 39.5 & 49.2 \\
FDA \cite{0001S20} & 92.5 & 53.3 & 82.4 & 26.5 & 27.6 & 36.4 & 40.6 & 38.9 & 82.3 & 39.8 & 78.0 & 62.6 & 34.4 & 84.9 & 34.1 & 53.1 & 16.9 & 27.7 & 46.4 &  50.5 \\
SA-I2I \cite{MustoZ20} & 91.2 & 43.3 & 85.2 & 38.6 & 25.9 & 34.7 & 41.3 & 41.0 & 85.5 & \bfseries46.0 & 86.5 & 61.7 & 33.8 & 85.5 & 34.4 & 48.7 & 0.0 & 36.1 & 37.8 & 50.4 \\
PIT \cite{LvLCL20} & 87.5  & 43.4  & 78.8  & 31.2  & 30.2  & 36.3  & 39.9  & 42.0  & 79.2  & 37.1  & 79.3  & 65.4  & 37.5  & 83.2  & 46.0  & 45.6  & 25.7  & 23.5  & 49.9 & 50.6 \\
IAST \cite{Mei_2020_ECCV} &  \bfseries93.8 &  57.8 & 85.1 & 39.5 & 26.7 & 26.2 & 43.1 & 34.7 & 84.9 & 32.9 &  88.0 & 62.6 & 29.0 & 87.3 & 39.2 & 49.6 & 23.2 & 34.7 & 39.6 & 51.5 \\
DACS \cite{Tranheden2021DACSDA}& 89.9 & 39.7 & \bfseries87.9& 30.7 & \bfseries39.5& 38.5 & 46.4 & \bfseries52.8 &88.0& 44.0 &\bfseries88.7& 67.0 &35.8 & 84.4& 45.7 & 50.2 & 0.0 & 27.2 & 34.0 & 52.1 \\
RPT$^\dagger$ \cite{zhang2020transferring} & 89.2 & 43.3 & 86.1 & 39.5 & 29.9 & 40.2 &  \bfseries 49.6 & 33.1 &  \bfseries87.4 & 38.5 & 86.0 & 64.4 & 25.1 &  88.5 & 36.6 & 45.8 &  23.9 & 36.5 &  \bfseries56.8 &  52.6 \\
    SAC\cite{DASAC} & 90.4 & 53.9 &  86.6 &  \bfseries42.4 & 27.3 &  45.1 & 48.5 &  42.7 &  \bfseries87.4 & 40.1 & 86.1 &  \bfseries67.5 & 29.7 & 88.5 &  49.1 &  \bfseries54.6 & 9.8 & 26.6 & 45.3 &  53.8 \\
    SAC* \cite{DASAC}& 89.9 &54.0&86.2&37.8&28.9& \bfseries45.9&46.9&47.7&88.0&44.8&85.5&66.4&30.3&\bfseries88.6&50.5&54.5&1.5&17.0&39.3&52.8\\
    
    \midrule
    DACS $+$ PAC (ours) &93.2&\bfseries58.8&87.2&33.3&35.1&38.6&41.8&51.4&\bfseries87.4&45.8&88.3&64.8&31.6&84.3&\bfseries51.7&53.4&0.6&31.3&50.6&\bfseries54.2\\
    \bottomrule
 \end{tabular}}
\end{table*}

%% file: sections/tab_syn2cscape.tex
\begin{table*}[h]
\caption{\footnotesize\textbf{\texttt{SYNTHIA $\rightarrow$ Cityscapes} results:}  Classwise and mean (over 16 classes) IoU comparison of our PAC-UDA with prior works. $^\dagger$ denotes the use of PSPNet \cite{ZhaoSQWJ17}, * denotes  our implementation of SAC with a restricted configuration (\texttt{GROUP\_SIZE}=2) compared to original SAC method (\texttt{GROUP\_SIZE}=4). All other methods use DeepLabV2 \cite{Chen2018DeepLabSI} architecture.}
\label{tab:s2c}
  \centering
  \scalebox{0.835}{
  \begin{tabular}{cccccccccccccccccc}
    \toprule
          & \rotatebox{90}{road}& \rotatebox{90}{sidewalk}& \rotatebox{90}{building} &\rotatebox{90}{wall} &\rotatebox{90}{fence} &\rotatebox{90}{pole} &\rotatebox{90}{light} & \rotatebox{90}{sign} &\rotatebox{90}{vegetable} &\rotatebox{90}{sky} &\rotatebox{90}{person} &\rotatebox{90}{rider} &\rotatebox{90}{car} &\rotatebox{90}{bus} &\rotatebox{90}{motor} &\rotatebox{90}{bike}  &mIoU\\
    \midrule
  SPIGAN\cite{Lee2019SPIGANPA} &71.1&29.8&71.4&3.7&0.3&33.2&6.4&15.6&81.2&78.9&52.7&13.1&75.9&25.5&10.0&20.5&36.8\\
   DCAN \cite{wu2018dcan}& 82.8& 36.4& 75.7& 5.1& 0.1& 25.8& 8.0& 18.7& 74.7& 76.9& 51.1& 15.9& 77.7& 24.8& 4.1& 37.3&  38.4\\ 
    DISE \cite{chang2019adapting}&  \bfseries91.7& \bfseries53.5& 77.1& 2.5& 0.2& 27.1& 6.2& 7.6& 78.4& 81.2& 55.8& 19.2& 82.3& 30.3& 17.1& 34.3&  41.5\\ 
    AdvEnt \cite{vu2018advent}& 85.6& 42.2& 79.7& 8.7& 0.4& 25.9& 5.4& 8.1& 80.4& 84.1& 57.9& 23.8& 73.3& 36.4& 14.2& 33.0&  41.2\\ DADA\cite{vu2019dada} &89.2&44.8&81.4&6.8&0.3&26.2&8.6&11.1&81.8&84.0&54.7&19.3&79.7&40.7&14.0&38.8&42.6\\
    CAG-UDA \cite{Zhang2019CategoryAU}& 84.7& 40.8& 81.7& 7.8& 0.0& 35.1& 13.3& 22.7& 84.5& 77.6& 64.2& 27.8& 80.9& 19.7& 22.7& 48.3& 44.5\\
PIT \cite{LvLCL20} & 83.1 & 27.6 & 81.5 & 8.9 & 0.3 & 21.8 & 26.4 &  \bfseries33.8 & 76.4 & 78.8 & 64.2 & 27.6 & 79.6 & 31.2 & 31.0 & 31.3 &  44.0 \\
PyCDA$^\dagger$ \cite{LianDLG19} & 75.5 & 30.9 & 83.3 & 20.8 & 0.7 & 32.7 & 27.3 & 33.5 & 84.7 & 85.0 & 64.1 & 25.4 & 85.0 & 45.2 & 21.2 & 32.0 &  46.7 \\
FADA \cite{Wang_2020_ECCV} & 84.5 & 40.1 & 83.1 & 4.8 & 0.0 & 34.3 & 20.1 & 27.2 & 84.8 & 84.0 & 53.5 & 22.6 & 85.4 & 43.7 & 26.8 & 27.8 &  45.2 \\
DACS\cite{Tranheden2021DACSDA} & 80.6 & 25.1 & 81.9 & 21.5& 2.9 & 37.2 & 22.7& 24.0 & 83.7 & \bfseries90.8& \bfseries67.6& \bfseries38.3& 82.9 & 38.9 & 28.5 & 47.6 & 48.3 \\
IAST \cite{Mei_2020_ECCV} & 81.9 & 41.5 & 83.3 & 17.7 & \bfseries 4.6 & 32.3 & 30.9 & 28.8 & 83.4 & 85.0 &  65.5 & 30.8 & 86.5 & 38.2 & \bfseries33.1 & 52.7 &  49.8 \\
RPT$^\dagger$ \cite{zhang2020transferring} & 88.9 & 46.5 & 84.5 & 15.1 & 0.5 & 38.5 & 39.5 & 30.1 & 85.9 & 85.8 & 59.8 & 26.1 & \bfseries 88.1 & \bfseries46.8 & 27.7 &  \bfseries56.1 &  51.2 \\ 
SAC \cite{DASAC}&  89.3 & 47.2 &  \bfseries85.5 &  26.5 & 1.3 &  \bfseries43.0 &  \bfseries45.5 & 32.0 &  \bfseries87.1 &  89.3 & 63.6 & 25.4 & 86.9 & 35.6 & 30.4 & 53.0 &  \bfseries52.6 \\
   SAC*\cite{DASAC}&\bfseries 91.7&52.7&85.1&22.6&1.5&42.2&44.1&30.9&82.5&73.8&63.0&20.9&84.9&29.5&26.9&52.2&50.3\\
   
     \midrule
      SAC $+$ PAC (ours) &83.2&40.5&85.4&\bfseries30.0&2.0&\bfseries43.0&42.2&\bfseries33.8&86.3&89.8&65.3&33.5&85.1&35.2&29.9&55.3&52.5\\
    \bottomrule
 \end{tabular}}
\end{table*}

%% file: sections/tab_ablations.tex
\begin{table}[h]
 \caption{\footnotesize\textbf{Ablations:} Comparing the effects of individual components of the regulariser (PAC) on final performance (mIoU). Here, the full model is DACS$+$PAC, and the adaptation setting is \texttt{GTA$\rightarrow$Cityscapes}; hyperparameters are: $k_s=50, b=200, \delta_\text{peak}=0.0025$; ``PL" refers to pseudo-labelling. We include classwise IoUs in the supplementary.}
  \centering
  \scalebox{1.21}{
  \begin{tabular}{cc}
    \toprule
     Configuration & mIoU \\
    \midrule
    All & 54.2 \\
      Only PL& 49.3\\
      Only Depth $+$ RGB segments & 51.9\\
      Only Depth segments w/ PL & 51.7\\
      Only RGB  segments w/ PL & 52.1\\

    \bottomrule

 \end{tabular}}
\label{tab:ablate}
\end{table}

%% file: sections/conclusion.tex
\section{Conclusion}
\label{sec:conclusion}
In this work, we proposed a multi-modal regularisation scheme for self-training approaches in unsupervised domain adaptation for semantic segmentation. We derive an objectness constraint from multi-modal clustering that is then used to formulate a contrastive objective for regularisation. We show that this regularisation consistently improves upon different types of self-training methods and even achieves state-of-the-art performance in popular benchmarks. In the future, we plan to study the effect of other modalities like 3D point-clouds in semantic segmentation.

%% file: sections/supplementary.tex
\appendix

\input{sections/algorithm_pac}
In this supplementary, we provide additional details and analysis for our proposed method, PAC-UDA. Algorithm \ref{alg:pac}, provides a step-by-step procedure for unsupervised domain adaptation via PAC-UDA. 

\section{Hyperparameters for Main Experiments}
\input{sections/tab_hypam}
To report the results in Table \ref{tab:generality}, Table \ref{tab:g2c} and Table \ref{tab:s2c}, we choose the best hyperparameters following standard cross-validation  on a random subset of Cityscapes train-split introduced in \cite{DASAC}. For base methods, we use the default hyperparameters from respective papers. In Table \ref{tab:hypam}, we summarise the hyperparameters for Table \ref{tab:generality}. Since the results of our approach in Table \ref{tab:g2c} and Table \ref{tab:s2c} are a subset of Table \ref{tab:generality}, the above hyperparameters apply there as well.

\section{Additional Ablations}
In this section, we provide additional ablation studies for DACS+PAC on \texttt{GTA$\rightarrow$Cityscapes}. The default hyperparameter configuration is: $k_s=25,~ b=200,~ \delta_\text{peak}=0.001, ~\tau_p=0.9$; unless otherwise stated. Also, we train each setting for $T_\text{train} (= 125\,000)$ iterations.

\subsection{Importance of Multiple Modalities and Pseudo-Labels}
\input{sections/tab_ablations_supplementary}
\begin{figure}[h]
\centering
 \includegraphics[width=0.9\linewidth]{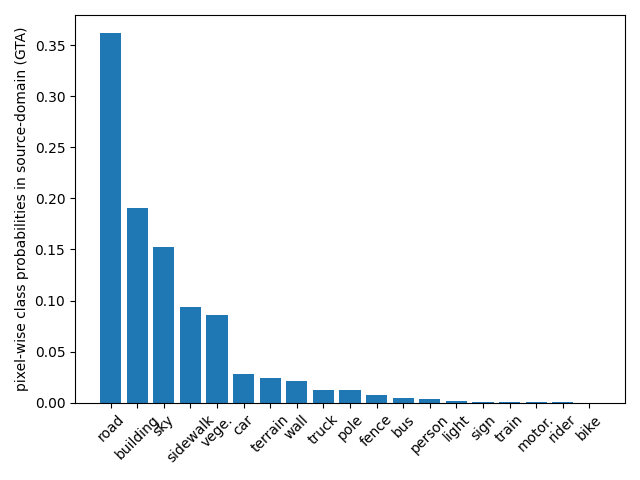}
\caption{Pixel-wise class distribution in GTA dataset}
\label{fig:class_dist}
\end{figure}

In Table \ref{tab:ablate_modality_full}, we provide the complete results (including classwise IoUs) for the ablations on individial modalities and pseudo-labels as described in Section \ref{sec:ablate}. While, Table \ref{tab:ablate_modality_full}  highlights the importance of combining all modalities with pseudo-labels for the best mean IoU, there are a few other important observations with respect to classwise IoUs. 

For instance, using ``PL'' for objectness constraints significantly \textit{underperforms} other settings (by upto $49$ IoU) in rare source-classes, like \textit{motorcycle} and \textit{bike} (Figure \ref{fig:class_dist}). This gap is surprisingly large (by upto $38.5$ IoU) even when compared to ``Depth-RGB''. We attribute this large performance gap to the class-imbalance problem \cite{zou2018unsupervised} that is known to adversely affect self-training in the absence of class-balanced losses. However, incorporating our objectness constraint alleviates the rare-class IoUs significantly without losing performance in frequent classes (except, \textit{sky}). These results provide strong evidence for the normalisation of class-related statistical effects in the presence of multimodal objectness-constraints.

Another interesting insight arises from comparing ``Depth-PL'' and ``RGB-PL'' settings that demonstrates the complementarity of the two modalities. Among the more frequent source-classes (Figure \ref{fig:class_dist}), purely RGB-based constraint considerably outperforms purely depth-based constraint in categories such as \textit{road}, \textit{sidewalk} and \textit{car} whereas the converse is true for other categories like \textit{wall}, \textit{pole}, \textit{terrain} and \textit{person}.  The outperformance of depth-based constraint on \textit{pole} and \textit{person} is intuitive since these objects have very small depth range compared to the scene depth and hence can be easily detected using the depth histogram (see Section \ref{sec:superv} for more discussion).

 \subsection{Importance of RGB-segments}
In the past, image clustering has been often used as an effective preprocessing step for segmentation \cite{zhang2020transferring, SLICsurvey}. Inspired by these works, in Table \ref{tab:ablate_slic}, we test the extent to which purely SLIC based RGB-segments can influence the objectness constraint and consequently, the final performance of our PAC-UDA. Specifically, we tabulate the performance with varying number of SLIC segments, $k_s$ and compare it to our default configuration, ``ALL ($k_s=25$)''.

We observe that when using only RGB-segments (without depth) for object-region estimates, there exists an intermediate value along the range of $k_s \in \{25, 50, 100\}$ where the semantic segmentation performance peaks. This trend empirically validates our intuition for choosing the best $k_s$ as discussed in Section \ref{sec:superv}. In fact, too small a value can be highly undesirable as it can lead to worse results ($52.1$ mIoU) than even the base method ($52.8$ mIoU). It is however, interesting to note that even with the most optimal $k_s=50$, just RGB based objectness constraint underperforms our multimodal constraint (``ALL'') by $\sim 1 $ mIoU.

 \subsection{Contrastive Loss Analysis}
\label{sec:ablate_contrast}
We analyze the effect of specific form of the contrastive loss function in Table \ref{tab:ablate_contrast}. Recall that in \eq \ref{eq:reguda}, our formulation of the contrastive loss maximizes the similarity of each pixel embedding, $\tilde{z}_p$ to \textit{only} a prototype of the region, $U_k$ that includes pixel $p$. Here, we introduce another variant of that loss, $L_{\text{obj}}^{t+}(p) $ that maximizes the similarity of $\tilde{z}_p$ to prototypes of \textit{all} valid regions, $\{U_k\}_{k=1}^K\setminus \Omega(k)$ that share the same region-label. While, originially, region-labels could influence the loss function only via dissimilarity scores, in $L_{\text{obj}}^{t+}(p) $, they can  influence  via both similarity and dissimilarity scores.

We observe that allowing greater region-label influence in $L_{\text{obj}}^{t+}(p) $ leads to worse mIoU than $L_{\text{obj}}^{t}(p) $. Zooming into the classwise IoUs reveal that less-common classes primarily contribute the the overall worse performance of  $L_{\text{obj}}^{t+}(p) $. We suspect that  increasing the influence of, and consequently the noise in, region-labels affect these less-common classes more adversely than common classes like \textit{road}, \textit{sidewalk}, \textit{wall} and \textit{car}. Finally, this ablation  guides our decision to adopt  $L_{\text{obj}}^{t}(p) $ as the default form of contrastive loss in \eq \ref{eq:reguda}. 

\begin{figure*}[!h]
\centering
 \includegraphics[width=\linewidth]{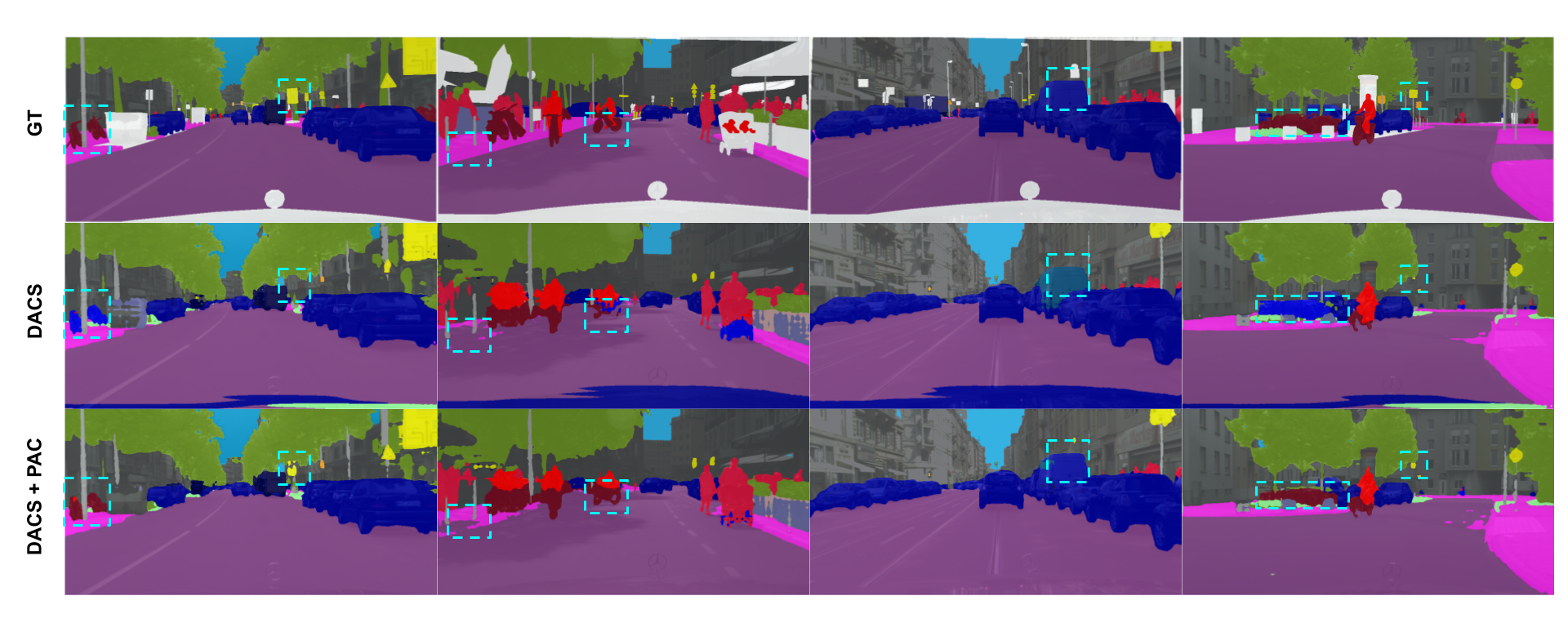}
\caption{\textbf{Additional qualitative results on Cityscapes\cite{cordts2016cityscapes} post adaptation from GTA\cite{richter2016playing}:} Blue dashed boxes highlight the semantic classes that our regularized version (DACS+PAC) is able to predict more reliably than the base method (DACS).}
\label{fig:qual_gta}
\end{figure*}

 \subsection{Importance of Region-Label Threshold}
An important hyperparameter of our objectness-constraint is the region-label threshold, $\tau_p$. At higher values of this threshold, valid object-regions are more likely to be a part of a single object and consistent with the ground-truth semantic category. This will positively influence the target-domain performance. At the same, time the number of such valid object-regions is likely to be small, which, may reduce the overall effect of the objectness-constraint on target-domain performance. As one decreases the threshold, the number of valid-regions will increase at the expense of region-label consistency with ground-truth. Thus, evaluating the performance over a range of values is crucial. 

Indeed, we observe in Table \ref{tab:ablate_thresh} that the mIoU increases with increase in threshold upto a certain point ($\tau_p=0.90$), beyond which the performance deteriorates. We, thus, set $0.90$ as our default threshold for all our experiments.

\section{Additional Visualisations}

In Figure \ref{fig:qual_gta}, we provide additional qualitative comparison between  DACS+PAC, DACS and the ground-truth under \texttt{GTA$\rightarrow$Cityscapes} settings.




%% file: sections/algorithm_pac.tex
\begin{algorithm*}[ht!]
  \caption{Unsupervised domain adaptation via PAC-UDA}
  \begin{algorithmic}[1]
  \Require{Pseudo-label ($\tilde{y}$); Target training dataset with depth ($D^t_\text{depth} = \{(x_i^t, h_i^t, \tilde{y}_i^t)\}_{i=1}^{N_{t}}$ ); Initial model parameters ($\theta_0 = \{\psi_0, \phi_0\}$); Number of histogram bins ($b$); Peak prominence threshold ($\delta_\text{peak}$); Number of RGB-segments ($k_s$); Spatial dimensions of depth map ($H\times W$); Region-label threshold ($\tau_p$); Objectness constraint loss weight ($\alpha_\text{obj}$);  Number of training iterations ($T_\text{train}$)}
 
 \Ensure{Target-domain adapted parameters ($\theta_* = \{\psi_*, \phi_*\}$)} 
\For{$t_\text{tr}$  $\gets 1~ \text{to} ~T_\text{train}$ }
\State $\{(x_i^t, h_i^t, y_i^t)\}_{i=1}^{N_t^B} \sim D_h^t$ \Comment{Randomly sample a training batch from target-domain}
\State Compute $L_\text{uda}$ \Comment{Self-training based adaptation objective (see Section \ref{sec:approach})}
\State $L_\text{obj} = 0$ \Comment{Initialise objectness-constraint}
\For{i  $\gets 1~ \text{to} ~N_{t}^{B}$ } 

 \State
\State Initialize $\mathcal{V}^d=\{\}$ \Comment{Empty list of \textbf{depth-segments}}
 \State $\texttt{Hist}\left(\{h_{im}\}_{m=1}^{HW};~b\right)$ $~\to~ \mathcal{F}^d$ \Comment{Histogram of depth values (HOD)}
 \State $\texttt{FindPeaks}$($ \mathcal{F}^d;~\delta_\text{peak}$) $~\to~ \{\mu_k\}_{k=1}^{k_d}$ \Comment{Cluster-center assignment using HOD}
\For{$k ~\gets~ 1~ \text{to} ~k_d$ } 
 \State $ V^d_k = \{m |m \in \{1,\ldots,HW\},~|h_m - \mu_k|<|h_m - \mu_{k'}|~ \forall k' \neq k \}$ \Comment{Depth segments}
 \State $\mathcal{V}^d.\texttt{append}(V_k^d)$ \Comment{Depth-segment list update}
 \EndFor
 \State
 
\State Initialize $\mathcal{V}^s=\{\}$ \Comment{Empty list of \textbf{RGB-segments}}
\State $\texttt{SLIC}$($ x_i;~k_s$) $~\to~ \{\mathcal{L}_k\}_{k=1}^{k_s}$ \Comment{RGB-segment labelling using $\texttt{SLIC}$\cite{6205760}}
\For{$k ~\gets~ 1~ \text{to} ~k_s$ } 
\State $V^s_k=\{m |m \in \{1,\ldots,HW\},~ \texttt{label}(m)=\mathcal{L}_k\}$ \Comment{RGB-segments}
 \State $\mathcal{V}^s.\texttt{append}(V_k^s)$ \Comment{RGB-segment list update}
 \EndFor
 \State

\State Initialize $\mathcal{V} =\{\}$ \Comment{Empty list of \textbf{object-regions}}
\State Initialize $k=0$ \Comment{region-index}
\For{$i'$  $\gets 1~ \text{to} ~k_s$ }  
\For{$j'$  $\gets 1~ \text{to} ~k_d$ }  
\State $k ~\gets~ k+1$ \Comment{Region-index update}
\State $V_{k}=\{m|m \in V_{i'}^s,~ m \in V_{j'}^d\}$ \Comment{Unique object-region assignment}
\State $\mathcal{V}.\texttt{append}(V_{k})$ \Comment{Object-region list update}
\EndFor 
\EndFor 
\State

\State $\mathcal{F}_k = \text{Histogram}(\{\tilde{y}^t_{im}\}_{m\in V_k}) ~\forall k=\{1, \ldots,K'\}$ \Comment{Region-wise frequency of pseudo-label classes}
\State Initialize $\mathcal{U} = \{\}$ \Comment{Empty list of valid regions}
\State Initialize $\mathcal{L} = \{\}$ \Comment{Empty list of valid region labels}
\For{$k ~\gets~ 1~ \text{to} ~K'$ } 
\If $\frac{\max \limits_c{\mathcal{F}_{k}[c]}}{\sum \limits_c\mathcal{F}_{k}[c]} \geq \tau_p$   \Comment{Threshold on majority-voting based region-label}
\State $U_k = V_k$ \Comment{Valid region assignment}
\State $\mathcal{U}.\texttt{append}(U_k)$ \Comment{Valid-region list update}
\State $\mathcal{L}_k = \arg \max \limits_{c} \mathcal{F}_k[c]$ \Comment{Region-label assignment}
\State $\mathcal{L}.\texttt{append}(L_k)$ \Comment{Valid-region label list update}
\EndIf
\EndFor   
\State Using $\mathcal{U}$ and $\mathcal{L}$, compute $L^t_{\text{obj}, i}$ \Comment{Objectness constraint, \eq. \ref{eq:reguda}} 
\State $L_\text{obj} = L_\text{obj} + L_{\text{obj},i}^t $ 
 \EndFor 
\State $L_\text{pac} = L_\text{uda}  + \frac{\alpha_\text{obj} }{N_t^B}* L_\text{obj}^t $ \Comment{Overall PAC-UDA objective}
\State $\theta_t ~\gets~ \theta_{t-1} - \eta \nabla L_\text{pac} $ \Comment{Parameter update}
\EndFor 
  \end{algorithmic}
  \label{alg:pac}
  \end{algorithm*} 
  

%% file: sections/tab_hypam.tex
\begin{table}[h]
 \caption{\footnotesize\textbf{Hyperparameters used in Table \ref{tab:generality}}}
  \centering
  \scalebox{0.9}{
  \begin{tabular}{ccccc}
    \toprule
     method & $k_S$ & $b$ & $\delta_\text{peak}$ & $\tau_p$ \\
    \midrule
    CAG + PAC & 50 & 200 & 0.0025 & 0.90 \\
      SAC + PAC& 50 & 200 & 0.0025 & 0.90\\
      DACS + PAC& 25 & 200 & 0.001 & 0.90\\

    \bottomrule
 \end{tabular}}
\label{tab:hypam}
\end{table}

%% file: sections/tab_ablations_supplementary.tex
\begin{table*}[h!]
 \caption{\footnotesize\textbf{Effect of Individual Modalities and Pseudo Labels:} Comparing the effects of individual modalities used to estimate object-regions and pseudo-labels on final performance (\textit{mIoU}). This table is an extended version of Table \ref{tab:ablate} with classwise IoUs. Mapping of configuration names to those in Table \ref{tab:ablate} - {\textbf{PL}}: Only PL; {\textbf{Depth-RGB}}: Only Depth+RGB segments; \textbf{Depth-PL}:  Only Depth segments w/ PL; {\textbf{RGB-PL}}: Only RGB segments w/ PL. Refer to Section \ref{sec:ablate} for configuration specific details.}
  \centering
  \scalebox{0.7}{
  \begin{tabular}{cccccccccccccccccccccc}
    \toprule
     Configuration & \rotatebox{90}{road}& \rotatebox{90}{sidewalk}& \rotatebox{90}{building} &\rotatebox{90}{wall} &\rotatebox{90}{fence} &\rotatebox{90}{pole} &\rotatebox{90}{light} & \rotatebox{90}{sign} &\rotatebox{90}{vege.} &\rotatebox{90}{terrain} &\rotatebox{90}{sky} &\rotatebox{90}{person} &\rotatebox{90}{rider} &\rotatebox{90}{car} &\rotatebox{90}{truck} &\rotatebox{90}{bus} &\rotatebox{90}{train} &\rotatebox{90}{motor} &\rotatebox{90}{bike} &\textbf{mIoU} \\
    \midrule   
	All & 93.2&58.8&87.2&33.3&35.1&38.6&41.8&51.4&87.4&45.8&88.3&64.8&31.6&84.3&51.7&53.4&0.6&31.3&50.6&54.2 \\
      PL& 93.7&58.7&86.8&27.3&29.7&35.4&41.6&50.6&87.1&46.7&89.2&65.2&37.1&87.4&41.3&49.8&0.0&7.0&1.6&49.3\\
      Depth-RGB &94.1&58.1&86.2&38.2&30.3&34.8&37.8&41.3&86.7&46.1&87.5&62.4&31.0&86.8&52.5&49.1&0.0&24.5&40.1& 51.9\\
     Depth-PL & 93.3&61.9&86.7&31.8&35.9&36.1&43.3&50.2&86.2&41.2&86.4&65.0&32.2&82.1&31.9&50.4&0.9&23.1&43.6& 51.7\\
      RGB-PL& 95.1&65.3&86.1&25.9&30.1&35.4&39.1&41.2&85.2&37.9&86.2&61.4&26.7&87.9&50.9&50.6&0.0&35.8&50.4& 52.1\\
    \bottomrule
 \end{tabular}}
\label{tab:ablate_modality_full}
\end{table*}

\begin{table*}[h!]
 \caption{\footnotesize\textbf{Importance of RGB-segments:} Comparing the effect of only RGB-segments with different  values of $k_s$. here, \textbf{PL}: Pseudo-Labels; \textbf{RGB-PL}: Objectness-constraint with only RGB segments and PL; \textbf{ALL}: Objectness-constraint with RGB-segments+Depth-segments and PL}
  \centering
  \scalebox{0.7}{
  \begin{tabular}{cccccccccccccccccccccc}
    \toprule
     Configuration & \rotatebox{90}{road}& \rotatebox{90}{sidewalk}& \rotatebox{90}{building} &\rotatebox{90}{wall} &\rotatebox{90}{fence} &\rotatebox{90}{pole} &\rotatebox{90}{light} & \rotatebox{90}{sign} &\rotatebox{90}{vege.} &\rotatebox{90}{terrain} &\rotatebox{90}{sky} &\rotatebox{90}{person} &\rotatebox{90}{rider} &\rotatebox{90}{car} &\rotatebox{90}{truck} &\rotatebox{90}{bus} &\rotatebox{90}{train} &\rotatebox{90}{motor} &\rotatebox{90}{bike} &\textbf{mIoU} \\
    \midrule   
 All ($k_s = 25$) & 93.2&58.8&87.2&33.3&35.1&38.6&41.8&51.4&87.4&45.8&88.3&64.8&31.6&84.3&51.7&53.4&0.6&31.3&50.6&54.2\\     
 RGB-PL($k_s = 25$) & 95.1&65.3&86.1&25.9&30.1&35.4&39.1&41.2&85.2&37.9&86.2&61.4&26.7&87.9&50.9&50.6&0.0&35.8&50.4& 52.1\\       
        RGB-PL ($k_s = 50$) &94.6&63.4&86.8&28.7&30.7&37.6&42.8&51.3&86.8&44.9&87.9&64.9&32.5&87.8&42.7&45.4&0.0&32.6&51.2&53.3\\
        RGB-PL ($k_s = 100$) &94.4&62.1&86.2&29.2&32.5&34.2&40.0&50.2&86.2&47.1&87.4&63.0&32.7&87.9&39.4&45.3&0.1&32.6&52.8&52.8 \\
         
    \bottomrule
 \end{tabular}}
\label{tab:ablate_slic}
\end{table*}

\begin{table*}[h!]
 \caption{\footnotesize\textbf{Effect of the Contrastive Objective:} Comparing two different formulations of contrastive objective as defined in \eq\ref{eq:reguda} and Section \ref{sec:ablate_contrast} and an upperbound configuration, GTlab (target-domain ground-truth labels}
  \centering
  \scalebox{0.73}{
  \begin{tabular}{cccccccccccccccccccccc}
    \toprule
     Configuration & \rotatebox{90}{road}& \rotatebox{90}{sidewalk}& \rotatebox{90}{building} &\rotatebox{90}{wall} &\rotatebox{90}{fence} &\rotatebox{90}{pole} &\rotatebox{90}{light} & \rotatebox{90}{sign} &\rotatebox{90}{vege.} &\rotatebox{90}{terrain} &\rotatebox{90}{sky} &\rotatebox{90}{person} &\rotatebox{90}{rider} &\rotatebox{90}{car} &\rotatebox{90}{truck} &\rotatebox{90}{bus} &\rotatebox{90}{train} &\rotatebox{90}{motor} &\rotatebox{90}{bike} &\textbf{mIoU} \\
    \midrule   
     $L_{\text{obj}}^{t}$ &93.2&58.8&87.2&33.3&35.1&38.6&41.8&51.4&87.4&45.8&88.3&64.8&31.6&84.3&51.7&53.4&0.6&31.3&50.6&54.2\\       
      $L_{\text{obj}}^{t+}$ & 94.2&59.4&86.7&35.8&32.1&36.8&40.5&49.4&86.5&41.9&86.0&63.5&27.1&89.1&53.7&54.5&2.5&27.3&45.7&53.3\\
    \bottomrule
 \end{tabular}}
\label{tab:ablate_contrast}
\end{table*}

\begin{table*}[h!]
 \caption{\footnotesize\textbf{Effect of region-label threshold, $\tau_p$ on Final Performance:} }
  \centering
  \scalebox{0.73}{
  \begin{tabular}{cccccccccccccccccccccc}
    \toprule
     Threshold & \rotatebox{90}{road}& \rotatebox{90}{sidewalk}& \rotatebox{90}{building} &\rotatebox{90}{wall} &\rotatebox{90}{fence} &\rotatebox{90}{pole} &\rotatebox{90}{light} & \rotatebox{90}{sign} &\rotatebox{90}{vege.} &\rotatebox{90}{terrain} &\rotatebox{90}{sky} &\rotatebox{90}{person} &\rotatebox{90}{rider} &\rotatebox{90}{car} &\rotatebox{90}{truck} &\rotatebox{90}{bus} &\rotatebox{90}{train} &\rotatebox{90}{motor} &\rotatebox{90}{bike} &\textbf{mIoU} \\
    \midrule   
      0.70 & 93.9&60.4&86.5&32.5&30.4&34.9&39.9&48.8&86.4&45.6&88.0&63.0&27.6&87.0&39.9&49.2&1.9&32.5&47.9&52.4 \\       
      0.80 &92.9&51.3&86.6&31.5&32.4&36.7&42.8&52.1&86.8&44.7&87.5&65.4&34.5&89.2&48.8&56.3&0.2&23.8&45.1&53.1\\
      0.90 &93.2&58.8&87.2&33.3&35.1&38.6&41.8&51.4&87.4&45.8&88.3&64.8&31.6&84.3&51.7&53.4&0.6&31.3&50.6&54.2\\
      0.95 &93.4&55.9&86.1&28.7&30.0&33.2&40.5&45.3&86.6&45.7&87.8&64.2&31.6&89.1&50.4&50.7&0.0&10.5&28.3&50.4\\
    \bottomrule
 \end{tabular}}
\label{tab:ablate_thresh}
\end{table*}